\documentclass[10pt,twocolumn,letterpaper]{article}

\usepackage{cvpr}              
\usepackage{multirow}
\usepackage[table]{xcolor}
\usepackage{colortbl}

\definecolor{cvprblue}{rgb}{0.21,0.49,0.74}
\usepackage[pagebackref,breaklinks,colorlinks,allcolors=cvprblue]{hyperref}

\def\paperID{} 
\def\confName{CVPR}
\def\confYear{2026}

\title{SDTrack: A Baseline for Event-based Tracking via Spiking Neural Networks}

\author{Yimeng Shan$^{1,2,3}$, Zhenbang Ren$^{1}$, Haodi Wu$^{1}$, 
Wenjie Wei$^{1}$, Rui-Jie Zhu$^{4}$, Shuai Wang$^{1}$, \\Dehao Zhang$^{1}$, Yichen Xiao$^{1}$, Jieyuan Zhang$^{1}$, Kexin Shi$^{1}$, Jingzhinan Wang$^{1}$, \\Jason K. Eshraghian$^{4}$, Haicheng Qu$^{3}$, Malu Zhang$^{\dag1,2}$\\
$^1$University of Electronic Science and Technology of China, China \\
$^2$Shenzhen Loop Area Institute, China\\
$^3$Liaoning Technical University, China\\
$^4$University of California, Santa Cruz, USA\\
\texttt{yimengshan2001@gmail.com} \\
Code: \url{https://github.com/YmShan/SDTrack}
}

\begin{document}
\maketitle

\begin{abstract}
Event cameras provide superior temporal resolution, dynamic range, energy efficiency, and pixel bandwidth. Spiking Neural Networks (SNNs) naturally complement event data through discrete spike signals, making them ideal for event-based tracking. However, current approaches combining Artificial Neural Networks (ANNs) and SNNs suffer from suboptimal architectures that compromise energy efficiency and limit tracking performance. To address these limitations, we propose the first Transformer-based \textbf{S}pike-\textbf{D}riven \textbf{T}racking (SDTrack) pipeline. It incorporates a novel event frame aggregation method called Global Trajectory Prompt (GTP) and a Transformer-based tracker. The GTP method effectively captures global trajectory information and aggregates it with event streams into event frames to enhance spatiotemporal representation. The Transformer-based tracker comprises a fully spike-driven SNN backbone and a simple tracking head. The SDTrack pipeline operates end-to-end without data augmentation or post-processing. Extensive experiments demonstrate that our SDTrack-Tiny pipeline achieves competitive accuracy with only 19.61$M$ parameters and 8.16$mJ$ energy consumption, while our Base version achieves state-of-the-art accuracy across three datasets. Our work establishes a solid foundation for future neuromorphic vision research.\end{abstract}

\section{Introduction}

\begin{figure}[ht]
  \centering 
  \includegraphics[width=0.45\textwidth]{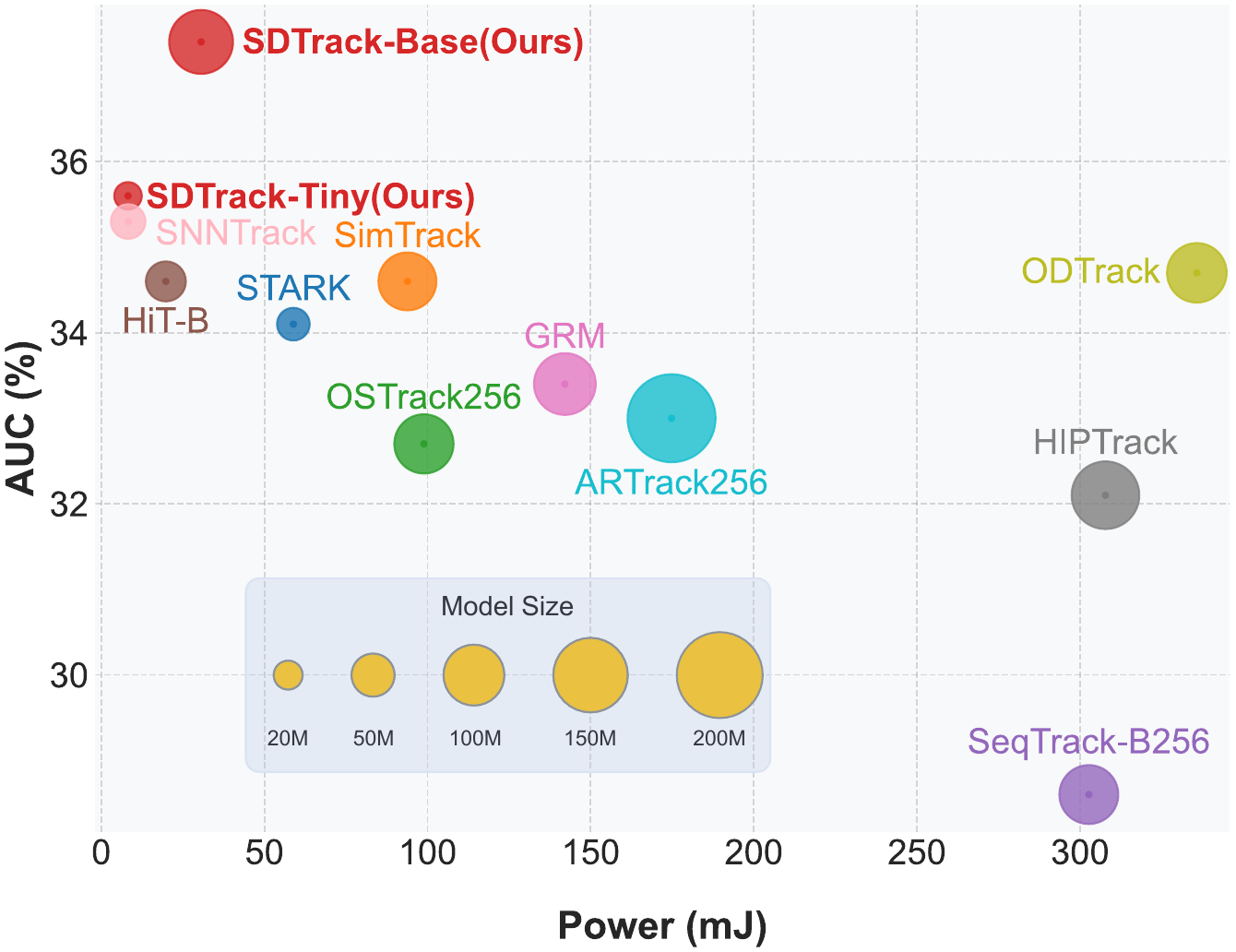}
  \caption{SDTrack versus other pipelines on VisEvent dataset: AUC, inference parameter count, and energy consumption.}
  \label{visual} 
\end{figure}

Visual object tracking is a fundamental task in computer vision, with extensive applications in surveillance, autonomous driving, and robots. Conventional RGB-based tracking methods have seen considerable advancements~\cite{chen2022backbone,chen2023seqtrack,chen2021transformer,yan2021learning,ye2022joint,zheng2024odtrack}, but often fail under challenging conditions such as low illumination~\cite{zou2012low}, overexposure~\cite{jatzkowski2018deep}, and high-speed motion~\cite{henriques2014high}.
In contrast, event cameras present attractive advantages, including higher temporal resolution (in the order of $\mu s$) and higher dynamic range (140dB vs. 60dB). These features enable event-based algorithms to locate objects in degraded conditions. 
Researchers have employed them for object tracking tasks. However, these works typically rely on Artificial Neural Networks (ANNs)~\cite{wang2023visevent,zhang2021object}, which completely ignores the sparse characteristics of event data and lead to high inference costs.

In contrast, Spiking Neural Networks (SNNs) naturally address this inefficiency. Compared with traditional ANNs, SNNs encode and propagate activation through sparse binary spikes, i.e., 0 or 1, which enables them to replace computationally intensive multiply-accumulate (MAC) operations in ANNs with energy-efficient accumulation (AC) operations. This characteristic makes SNNs naturally align with the sparse feature of event data and the low power consumption properties of event cameras~\cite{wei2026tp,wang2025spiking,liu2026hardf}. Consequently, the integration of SNNs into event-based tracking emerges as a compelling and synergistic approach.

Several studies have applied SNNs to event-based tracking with promising results. However, most of these works adopt a hybrid architecture that combines ANNs and SNNs~\cite{zhang2021object,zhang2025spiking}, which cannot fully exploit the power-efficient nature of SNNs. Moreover, they do not utilize cross-correlation operations between template and search features for relational modeling, which has been demonstrated to improve tracking performance~\cite{ye2022joint}.
To address these limitations, we propose the first transformer-based \textbf{S}pike-\textbf{D}riven \textbf{T}racking (SDTrack) Pipeline, which achieves state-of-the-art (SOTA) performance in event-based tracking without requiring data augmentation or post-processing.

For the SDTrack Pipeline's input, we observe that existing event aggregation methods exhibit insufficient capacity to generate robust representations for real-time event flow.
To address this, we propose a Global Trajectory Prompt (GTP) method  that effectively captures trajectory information within the event flow.
Specifically, GTP accumulates positive and negative polarities in the first two channels respectively, while recording trajectory information in the third channel. 
This three-channel frame representation of event data provides enhanced temporal information representation and effectively aligns with data formats commonly used in computer vision tasks, facilitating better utilization of pre-trained weights for training the tracker.

For the tracker in the SDTrack Pipeline, we primarily conduct designs in two aspects, namely position information acquisition and backbone architecture. First, we design a method named Intrinsic Position Learning (IPL) that enables the network to autonomously learn position information without introducing additional parameters, which also allows the tracker to learn stronger semantic information of the target to be tracked. Then, for the backbone, we analyze the problems existing in several candidate backbones and design a more appropriate SNN-based Transformer network as the backbone for the tracking task.

We conduct extensive experiments to access our methods. On one hand, we validate the efficacy and generality of our GTP method, providing new baselines for mainstream trackers across multiple event-based benchmarks.
On the other hand, we demonstrate that SDTrack achieves an excellent balance among accuracy, parameter count, and energy consumption, as depicted in Fig.~\ref{visual}.
The main contributions of this paper are summarized as follows:
\begin{itemize}
    \item We propose a novel event aggregation method called GTP, which effectively captures global trajectory information to yield enhanced spatiotemporal representations and contributes higher accuracy for various trackers.

    \item We propose a tracker incorporating two innovations: the IPL method that enhances position learning without additional parameters, and an SNN-based Transformer backbone tailored for tracking tasks.

    \item Based on the GTP method and our proposed tracker, we construct SDTrack, the first fully SNN-based event-based tracking pipeline that operates end-to-end without data augmentation or post-processing.

    \item SDTrack achieves competitive performance on multiple event-based tracking benchmarks with the lowest parameter count and energy consumption. These results demonstrate the potential of SNNs for event-based tracking and establish a solid baseline for future research.

\end{itemize}

\section{Related Work}

\textbf{Event Aggregation Method in Tracking.} To leverage advanced trackers designed for traditional cameras, researchers typically divide asynchronous event flow into sub-event streams of equal length and aggregate them into synchronous event frames. However, current event aggregation methods in the tracking domain are suboptimal. The Event Frame method only records the latest polarity at each spatial location~\cite{zhang2022spiking,wang2023visevent}, losing valuable information when multiple changes occur at a pixel. While Time-Surface~\cite{lagorce2016hots} and Event Count~\cite{maqueda2018event} methods can implicitly represent multi-directional motion, their temporal representation lacks robust trajectory information for tracking. Zhu et al.\cite{zhu2018ev} and Wang et al.\cite{wang2019ev} aggregate latest occurrence times of polarity changes, but their four-channel representation performs poorly in complex tasks and limits transfer learning, as pre-trained weights are designed for three-channel inputs~\cite{he2022masked,he2016deep}. Therefore, we propose GTP, which captures both short-term multi-directional movements and global trajectory information, enhancing spatiotemporal representation while maximizing transfer learning potential.

\begin{figure*}[ht]
  \centering 
  \includegraphics[width=1.0\textwidth]{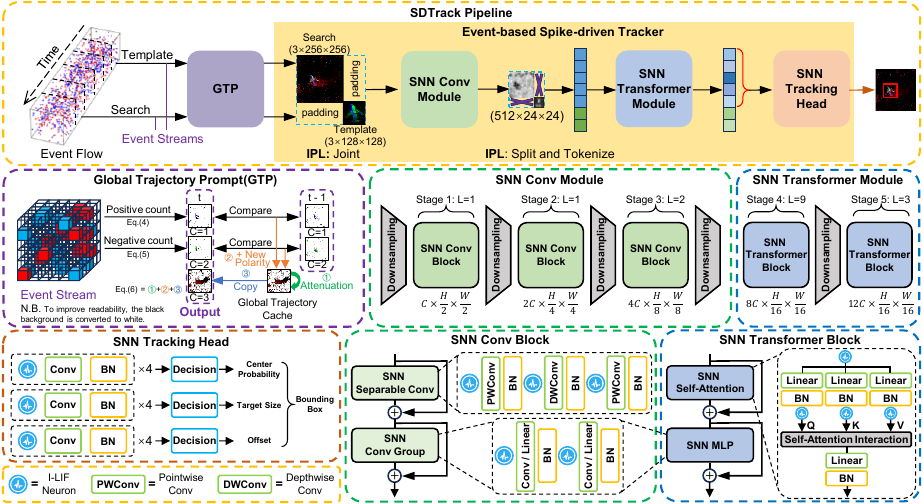}
  \caption{Overview of the SDTrack Pipeline. Upon receiving template and search event streams, GTP aggregates them into event frames, which are then concatenated into a unified matrix by IPL. Following SNN Conv Block processing, the matrix undergoes restoration and tokenization. The template features, after cross-correlation with search features, are fed into the SNN Tracking Head for target position and scale prediction. The detailed design of each module is illustrated in the middle and bottom panels.}
  \label{main} 
\end{figure*}

\textbf{Event-based Tracking.} Various studies demonstrate that trackers in RGB-based single object tracking (SOT) perform inadequately on event-based tracking tasks~\cite{wang2024long,wang2023visevent}. Research indicates this deficiency stems from the lack of texture information needed by RGB-based trackers in the event data~\cite{zhang2022spiking}.
Given the sparse spike-driven nature of SNNs, several studies suggest an inherent compatibility between SNNs and event data~\cite{steffen2019neuromorphic,chaney2021self}. Consequently, studies have employed SNNs to construct event-based trackers~\cite{debat2021event,nagaraj2022dotie,kosta2025toffe} or tracking systems for edge devices~\cite{liu2022real,huang20231000}.
Among these SNN-based trackers, STNet~\cite{zhang2022spiking} and SNNTrack~\cite{zhang2025spiking} that utilize SNNs to extract temporal features have emerged as competitive trackers. 
However, both of them implement hybrid architectures combining SNNs and ANNs, and neither leverages self-attention for template-search interaction. 
This results in the energy efficiency benefit and performance potential of SNNs still not being fully utilized.
To address these limitations, we design the first transformer-based spike-driven tracker, aimed at leveraging the inherent energy efficiency advantage of SNNs while fully unleashing their performance potential.

\section{SDTrack Pipeline}

The SDTrack Pipeline comprises a \textbf{GTP method} for robust spatiotemporal event representation and a \textbf{spike-driven tracker}, as shown in Fig.~\ref{main}. This section details the key architectural components of the system: the Spiking Neuron, GTP method implementation, and tracker elements including IPL method, backbone, and tracking head.

\subsection{Preliminary}
\label{sec:ILIF}

The brain's neural system serves as the source of neural network innovation. Its diverse neural dynamics inspire researchers to design various types of spiking neurons for SNNs. However, \textbf{our method is general and does not rely on any specific neuron type}. Therefore, we present a unified dynamical equation for spiking neurons, described as
\begin{align}
\mathbf{U}[t] &= \mathbf{H}[t - 1] + \frac{1}{\tau }(\mathbf{X}[t] - (\mathbf{H}[t - 1] - {\mathbf{U}_{rest}})), \label{eq1} \\
\mathbf{S}[t] &= f(\mathbf{U}[t] - {\mathbf{U}_{thr}}), \label{eq2} \\
\mathbf{H}[t] &= \mathbf{U}[t](1 - \mathbf{S}[t]).  \label{eq3}
\end{align}

For any neuron, when it receives input $\mathbf{X}[t]$ at time $t$, the membrane potential rises to $\mathbf{U}[t]$. Eq.~\ref{eq2} then determines the spike value at time $t$, where $\mathbf{S}[t]=0$ indicates no spike emission, and $\mathbf{U}_{thr}$ represents the threshold. $\mathbf{H}[t]$ denotes the membrane potential after completing Eq.~\ref{eq2}. $\tau$ serves as the decay factor for the neuron's membrane potential, simulating the stimulus decay when the neuron receives no stimulation for extended periods. $\mathbf{U}_{rest}$ represents the reset membrane potential after neuronal discharge.

Mainstream spiking neuron models can be represented by Eq.\ref{eq1}-\ref{eq3}. For instance, when function $f( \cdot )$ is the Heaviside step function, if $\tau$ equals 1, the model represents an IF neuron~\cite{brunel2007lapicque}, while when $\tau$ exceeds 1, it represents an LIF neuron~\cite{maass1997networks}. When $\mathbf{U}_{rest}=\mathbf{U}_{thr}=0$ and $f(x)={\frac{1}{D} \cdot \operatorname{Clip}(\operatorname{round}(x),0,D)}$, the model becomes an I-LIF model~\cite{yao2024scaling}, where $x=\mathbf{U}[t]$ and $D$ represents virtual time steps. $\operatorname{clip}(x, min, max)$ constrains $x$ to $[min, max]$, and $\operatorname{round}(\cdot)$ performs rounding. During inference, this model can be converted to 0/1 spike signals using the spike-ahead principle, where $D$ is incorporated into the actual iterative timestep $T$, forming a total of $T \times D$.

Therefore, these neurons achieve spike-driven inference, replacing MAC operations with AC operations, thereby significantly reducing computational costs.

\begin{figure}[ht]
  \centering 
  \includegraphics[width=0.43\textwidth]{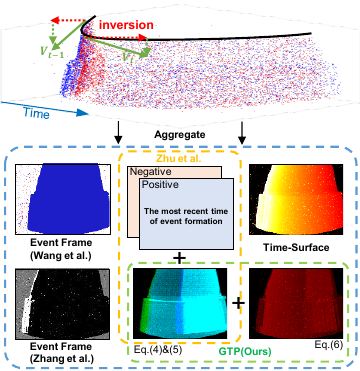}
  \caption{Example of event stream and event aggregation. The upper shows an event stream of a suspended object moving rightward then leftward from center in a short time. The lower displays results using various event aggregation methods for this stream. }
  \label{data_aggregate} 
\end{figure}

\subsection{Global Trajectory Prompt}
\label{sec:GTP}

Event cameras respond to brightness changes asynchronously and independently for each pixel. For each pixel, when the change in log-scale intensity exceeds a threshold, it outputs an event $e=\{x, y, t, p\}$, where $x$ and $y$ are spatial coordinates, $t$ is the timestamp, and $p$ is the polarity.
To leverage rich research achievements in vision models, many studies formally aggregate event streams into event frames and then process them using vision models. These methods discard the unique spatiotemporal characteristics of event data and cannot fully utilize the potential of pre-trained vision models. Our proposed GTP method preserves strong spatiotemporal information through accumulation and trajectory representations, while the three-channel event frame representation ensures that the potential of pre-trained vision models is fully released, effectively addressing the aforementioned issues.
Specifically, for the $i$-th event frame, GTP computes pixel-wise accumulation in the first two channels using the following equations
\begin{align}
\label{eq4}
h_i^1(x,y) \buildrel\textstyle.\over= \alpha  \cdot \sum\limits_{{t_k} \in L} {\delta (x - {x_k},y - {y_k})\delta ({p_k} - 1)},\\
h_i^2(x,y) \buildrel\textstyle.\over= \alpha  \cdot \sum\limits_{{t_k} \in L} {\delta (x - {x_k},y - {y_k})\delta ({p_k} + 1)},
\end{align}
where $L$ is the duration of each event stream and $\alpha$ is a coefficient that preserves more valid data while removing partial noise.
By using our GTP, the first and second channels in a event frame, i.e., $h_i^1$ and $h_i^2$, accumulate the number of positive and negative polarities within the event stream, respectively.
As shown in Fig.~\ref{data_aggregate}, when multiple directional movements occur at a pixel within the interval $L$, prior methods only select the most recent polarity value as the aggregation result, discarding preceding motion information. In contrast, GTP separately accumulates the number of positive and negative polarities, thus preserving all motion information throughout the entire time interval.

In addition to recording all motion information via the first two channels, GTP captures global trajectory information in the third channel of the event frame, described as
\begin{equation}
\label{eq4}
h_i^3(x,y) \buildrel\textstyle.\over= h_{i - 1}^3(x,y) \cdot \beta  + \alpha  \cdot \sum\limits_{j = 1}^2 {C(h_{i - 1}^j(x,y),h_i^j(x,y))},
\end{equation}

\begin{equation}
\label{eq5}
C(h_{i - 1}^j, h_i^j) = \mathbb{I}(h_{i - 1}^j = 0 \text{ and } h_i^j \neq 0),
\end{equation}
where $\mathbb{I}$ is the Boolean indicator function, $\delta$ denotes the Kronecker delta function, and $\beta$ is a decay factor. $h_0^3$ is initialized as a zero matrix. When accumulating the third channel of the i-th event frame, we decay each pixel in the previous event frame $h_{i-1}^3$ by factor $\beta$, serving as initial information for the current third channel.
The third channel aggregation completes after the first two channels. As shown in Fig.~\ref{data_aggregate}, the GTP method captures global trajectory information embedded within event data, thereby providing cues for object position and shape in tracking.

Overall, the $i$-th event frame is constructed by stacking $h_i^1$, $h_i^2$, and $h_i^3$. This three-channel representation exhibits stronger compatibility with existing computer vision architectures, thereby enabling the tracker to better acquire feature extraction and perception capabilities from visual pre-training. Subsequently, the event-based tracker decouples the strong spatial information from the first two channels through temporal information provided by the third channel to predict textures, thus precisely capturing target locations.

\subsection{Transformer-based Spike-driven Tracker}
\label{sec:SDTrack}

The tracker in the SDTrack Pipeline is illustrated in the yellow of Fig.~\ref{main}. This section describes its design details.

\textbf{Intrinsic Position Learning.} Motivated by two key observations. First, joint positional encoding of template and search frames enhances their cross-correlation performance~\cite{kang2023exploring}. Second, residual convolutional blocks preceding SNN-based Transformer architectures enable effective learning of positional information~\cite{zhou2022spikformer,yao2024spikev1,yao2024spikev2}. Based on these insights, we propose Intrinsic Position Learning (IPL). In SDTrack Pipeline, IPL concatenate the event-aggregated template frame $\mathbf{Z} \in (T,C,{H_z},{W_z})$ and search frame $\mathbf{X} \in (T,C,{H_x},{W_x})$ diagonally before feeding them into the SNN Conv Module
\begin{equation}
\mathbf{U} = \operatorname{IPL}(\mathbf{X},\mathbf{Z}),
\end{equation}
\begin{equation}
\label{eq6}
\operatorname{IPL}(\mathbf{X},\mathbf{Z}) = \left[ {\begin{array}{*{20}{c}}
\mathbf{X}&{{{\rm O}_1}}\\
{{{\rm O}_2}}&\mathbf{Z}
\end{array}} \right],
\end{equation}
where $\mathbf{U} \in (T,C,{H_z+H_x},{W_z+W_x})$, ${{\rm O}_1}$ and ${{\rm O}_2}$ represent zero matrixs. As demonstrated in Tab.~\ref{ablation}, our approach enables SDTrack to capture stronger positional information compared to conventional encoding schemes, without explicitly requiring positional encoding. Due to the spike-driven nature of SNNs, the additional padding introduces negligible computational overhead.

\textbf{Backbone.} The tracker backbone comprises an SNN Conv Module with several SNN Conv Blocks and an SNN Transformer Module with several SNN Transformer Blocks. The SNN Conv Block can be written as
\begin{align}
  &\quad \mathbf{U}' = \mathbf{U} + \operatorname{SNNSepConv}(\mathbf{U}), \\
  &\quad \mathbf{U}'' = \mathbf{U}' + \operatorname{SNNConvGroup}(\mathbf{U}'), \\
  &\quad \operatorname{SNNConvGroup}(\mathbf{U}')= f(f(\mathbf{U}')), \\
  &\text{where } \operatorname{SNNSepConv}(\mathbf{U}) \text{ is computed as:} \notag \\
  &\quad \operatorname{SNNSepConv}(\mathbf{U}) = \operatorname{Conv_{pw}}(\mathcal{SN}(\mathbf{V}_2)), \\
&\quad \mathbf{V}_2 = \operatorname{Conv_{dw}}(\mathcal{SN}(\mathbf{V}_1)), \\
  &\quad \mathbf{V}_1 = \operatorname{Conv_{pw}}(\mathcal{SN}(\mathbf{U})), 
\end{align}
where $\mathbf{U}$ represents the input membrane potential, $\mathbf{U}'$, $\mathbf{U}''$, $\mathbf{V}_1$, and $\mathbf{V}_2$ denote the intermediate membrane potential variables illustrated in the middle section of Fig.~\ref{main}, $\mathcal{SN}(\cdot)$ is the spiking neuron layer, $\operatorname{Conv_{pw}}(\cdot)$ and $\operatorname{Conv_{dw}}(\cdot)$ are pointwise convilution and depthwise convolution. $f$ denotes the $\operatorname{Conv(\mathcal{SN}(\cdot))}$ operation. Note, we ignore the batch normalization layer for ease of writing.

Before feeding them into the SNN Transformer Module, they are split back to their original positions prior to entering the SNN Conv Module and tokenized. The SNN Transformer Block is formulated as follows
\begin{align}
&\mathbf{U}' = \mathbf{U} + \operatorname{SSA}(\mathbf{U}), \\
&\mathbf{U}'' = \mathbf{U}' + \operatorname{SNNMLP}(\mathbf{U}').
\end{align}

The SNN Self-Attention (SSA) module enables tokens with high-level features from both template and search regions to interact through correlation operations, extracting target-oriented features. The input token sequence is transformed into Query ($\mathbf{Q_s}$), Key ($\mathbf{K_s}$), and Value ($\mathbf{V_s}$) spikes through three learnable linear matrices. The correlation operation process can be described as
\begin{equation}
\operatorname{SSA}(\mathbf{Q_s},\mathbf{K_s},\mathbf{V_s}) = \mathbf{Q_s}{\mathbf{K_s}^\top}\mathbf{V_s} * s,
\end{equation}
where $s$ is a scaling factor that is set in relation to the input channel dimensions and the number of attention heads.

\textbf{Tracking Head.} We construct an SNN version of the center prediction head~\cite{ye2022joint} to build the tracker, which localizes targets by predicting their center positions and dimensions. We find that in the SDTrack pipeline, the center head outperforms the corner head, as shown in Tab.~\ref{abla}.

\textbf{Training and Inference.} Similar to prior works, we train SDTrack through frame pair matching~\cite{ye2022joint,chen2022backbone,yan2021learning}. The loss computation incorporates both classification and regression losses. We adopt the weighted focal loss~\cite{law2018cornernet} for classification. With the predicted bounding box, $\mathcal{L}1$ loss and the generalized IoU loss~\cite{rezatofighi2019generalized} are employed for bounding box regression. The overall loss function is
\begin{equation}
\mathcal{L} = {\mathcal{L}_{\operatorname{cls}}} + {\lambda _{\operatorname{iou}}}{\mathcal{L}_{\operatorname{iou}}} + {\lambda _{\mathcal{L}1}}{\mathcal{L}_1},
\end{equation}
where ${\lambda _{iou}}=2$ and ${\lambda _{L1}}=5$ are the regularization parameters following~\cite{yan2021learning}.

During inference, we follow the standard SOT procedure using the first frame of the sequence as template~\cite{ye2022joint}. Notably, we do not employ dynamic template strategies or post-processing techniques such as hanning window penalty. Our GTP method is applicable to real-world scenarios in real-time, as illustrated by the purple box in Fig.~\ref{main}.

\begin{table*}[!htbp]
\centering
\small
\begin{tabular}{@{}ccccccccccc@{}}
\toprule
\multirow{2}{*}{Methods}  
  & \multirow{2}{*}{\begin{tabular}[c]{@{}c@{}}Param.\\ (M)\end{tabular}} 
  & \multirow{2}{*}{\begin{tabular}[c]{@{}c@{}}Spiking \\ Neuron\end{tabular}} 
  & \multirow{2}{*}{\begin{tabular}[c]{@{}c@{}}Timesteps\\($T\times D$)\end{tabular}} 
  & \multirow{2}{*}{\begin{tabular}[c]{@{}c@{}}Power\\ (mJ)\end{tabular}} 
  & \multicolumn{2}{c}{FE108} 
  & \multicolumn{2}{c}{FELT} 
  & \multicolumn{2}{c}{VisEvent} 
\\  
\cmidrule(l){6-11} 
  & & & & 
    & AUC(\%) & PR(\%) 
    & AUC(\%) & PR(\%) 
    & AUC(\%) & PR(\%) 
\\ \midrule
STARK~\cite{yan2021learning}          & 28.23 & –    & $1\times1$ & 58.88 & 57.4 & 89.2 & *39.3 & *50.8 & 34.1 & 46.8 \\
SimTrack~\cite{chen2022backbone}     & 88.64 & –    & $1\times1$ & 93.84 & 56.7 & 88.3 & 36.8  & 47.0  & 34.6 & 47.6 \\
OSTrack\textsubscript{256}~\cite{ye2022joint}       & 92.52 & –    & $1\times1$ & 98.90 & 54.6 & 87.1 & 35.9  & 45.5  & 32.7 & 46.4 \\
ARTrack\textsubscript{256}~\cite{wei2023autoregressive}
                                      &202.56 & –    & $1\times1$ &174.80 & 56.6 & 88.5 & \textcolor{blue}{\textbf{39.5}} & 49.4 & 33.0 & 43.8 \\
SeqTrack-B\textsubscript{256}~\cite{chen2023seqtrack}
                                      & 90.60 & –    & $1\times1$ &302.68 & 53.5 & 85.5 & 33.0  & 42.0  & 28.6 & 43.3 \\
HiT-B~\cite{kang2023exploring}        & 42.22 & –    & $1\times1$ & 19.78 & 55.9 & 88.5 & 38.5  & 48.9  & 34.6 & 47.6 \\
GRM~\cite{gao2023generalized}         & 99.83 & –    & $1\times1$ &142.14 & 56.8 & 89.3 & 37.2  & 47.4  & 33.4 & 47.7 \\
HIPTrack~\cite{cai2024hiptrack}       &120.41 & –    & $1\times1$ &307.74 & 50.8 & 81.0 & 38.2  & 48.9  & 32.1 & 45.2 \\
ODTrack~\cite{zheng2024odtrack}       & 92.83 & –    & $1\times1$ &335.80 & 43.2 & 69.7 & 29.7  & 35.9  & 24.7 & 34.7 \\
*SiamRPN~\cite{li2018high}            &   –   & –    & $1\times1$ &   –   & –    & –    & –     & –     & 24.7 & 38.4 \\
*ATOM~\cite{danelljan2019atom}        &   –   & –    & $1\times1$ &   –   & –    & –    & 22.3  & 28.4  & 28.6 & 47.4 \\
*DiMP~\cite{bhat2019learning}         &   –   & –    & $1\times1$ &   –   & –    & –    & 37.8  & 48.5  & 31.5 & 44.2 \\
*PrDiMP~\cite{danelljan2020probabilistic} 
                                      &   –   & –    & $1\times1$ &   –   & –    & –    & 34.9  & 44.5  & 32.2 & 46.9 \\
*MixFormer~\cite{cui2022mixformer}     & 37.55 & –    & $1\times1$ &   –   & –    & –    & 38.9  & 50.4  & –    & –    \\
*STNet~\cite{zhang2022spiking}        &\textcolor{blue}{\textbf{20.55}} & LIF & $3\times1$ &   103.53   & –    & –    & –     & –     & 35.0 & 50.3 \\
*SNNTrack~\cite{zhang2025spiking}     & 31.40 & BA-LIF & $5\times1$ & 8.25  & –    & –    & –     & –     & 35.4 & \textcolor{blue}{\textbf{50.4}} \\  
\midrule
\multirow{3}{*}{\textbf{SDTrack-Tiny}} 
  & \multirow{3}{*}{\textcolor{red}{\textbf{19.61}}} & LIF    & $4\times1$ & \textcolor{red}{\textbf{8.15}} 
    & 56.7 & 89.1 & 35.8 & 44.0 & 35.4 & 48.7 \\
  & & I-LIF  & $2\times2$ & 9.87 
    & 55.3 & 88.1 & 35.7 & 45.3 & 35.4 & 49.5 \\
  & & I-LIF  & $1\times4$ & \textcolor{blue}{\textbf{8.16}}
    & \textcolor{blue}{\textbf{59.0}} & \textcolor{blue}{\textbf{91.3}} & 39.3 & \textcolor{blue}{\textbf{51.2}} & \textcolor{blue}{\textbf{35.6}} & 49.2 \\ 
\cmidrule(l){2-11}
\textbf{SDTrack-Base} 
  & 107.26 & I-LIF & $1\times4$ & 30.52 
    & \textcolor{red}{\textbf{59.9}} & \textcolor{red}{\textbf{91.5}} & \textcolor{red}{\textbf{40.0}} & \textcolor{red}{\textbf{51.4}} & \textcolor{red}{\textbf{37.4}} & \textcolor{red}{\textbf{51.5}} \\ 
\bottomrule
\end{tabular}

\caption{Comparison with standard SOT pipelines on three event-based tracking benchmarks. * denotes results directly adopted from their respective benchmark reports. Energy consumption is measured on the VisEvent dataset. The best two results are shown in \textcolor{red}{\textbf{red}} and \textcolor{blue}{\textbf{blue}}.}
\label{main_table}
\end{table*}

\section{Experiments}

\subsection{Implement Details}

The Base-scale tracker architecture (Fig.~\ref{main}, middle) sets $C = 64$.
The Tiny-scale tracker sets $C = 32$, reduces Stage 4 to 6 blocks, and employs 360 channels with 2 blocks in Stage 5, following~\cite{yao2024scaling}.
The tracking heads differ only in channel dimensions: the first convolutional layer outputs 256 channels in the Tiny model versus 512 in the Base, with subsequent layers progressively halved.
Both models maintain input dimensions of $128 \times 128$ and $256 \times 256$ for template and search frames, respectively.
When $T > 1$, the event stream is divided into $T$ portions to execute GTP.

We first pre-train the backbone of SDTrack on the ImageNet-1K dataset. We then fine-tune it on the corresponding event-based tracking datasets through a pair matching task~\cite{yan2021learning,chen2022backbone,ye2022joint,kang2023exploring}. During this process, no data augmentation or preprocessing is utilized. The hyperparameters for the fine-tuning process depend on the dataset and model size. Complete details of pre-training and fine-tuning are available in the Supplementary Materials. All experiments are conducted on 4 * NVIDIA RTX 4090 GPUs. We will release all available models and code.

\subsection{Evaluation Metric}
We conduct comparative analyses with existing methods primarily across four metrics: Area Under the Curve (AUC), Precision Rate (PR), parameter count, and theoretical energy consumption. The AUC is computed from the area under the Success Plot curve, while PR represents the percentage of samples where the distance between the predicted and ground truth target centers is below 20 pixels.
Following energy estimation approaches in prior works~\cite{yao2024spikev1,yao2024spikev2,yao2024scaling}, the theoretical energy consumption per inference of the tracker of SDTrack can be calculated as
\begin{align}
E &= T \cdot ( {E_{\operatorname{MAC}}} \cdot \sum\limits_{i} {\operatorname{FL}_{\operatorname{SNNConv}}^i} \\& + {E_{\operatorname{AC}}} \cdot 
(\sum\limits_{j} {\operatorname{FL}_{\operatorname{SNNConv\&FC}}^j \cdot f{r^j}) + \sum\limits_{k} {\operatorname{FL}_{\operatorname{SSA}}^k }} )\nonumber,
\end{align}
where ${E_{\operatorname{MAC}}}$ and ${E_{\operatorname{AC}}}$ denote the energy costs of MAC and AC operations, respectively, while $\operatorname{FL}_{\operatorname{SNNConv\&FC}}$ represents the FLOPs of the $j$-th convolutional layer or FC layer, and $\operatorname{FL}_{\operatorname{SSA}}^k$ represents the FLOPs of the $k$-th SNN Self-Attention module. Additionally, ${f{r^j}}$ denote the firing rate (the proportion of non-zero elements) of the corresponding spike matrices at matching positions. Following standard practice~\cite{yao2024spikev1,yao2024spikev2,yao2024scaling}, we assume the data for various operations is implemented using 32-bit floating-point precision in 45nm technology~\cite{horowitz20141}, where ${E_{\operatorname{MAC}}} = 4.6pJ$ and ${E_{\operatorname{AC}}} = 0.9pJ$. The detailed computation methods for all metrics and the neuron firing rates in the in tracker of SDTrack are presented in the Supplementary Materials.

\subsection{Comparison with Standard Pipeline}
\begin{figure}[ht]
  \centering 
  \includegraphics[width=0.47\textwidth]{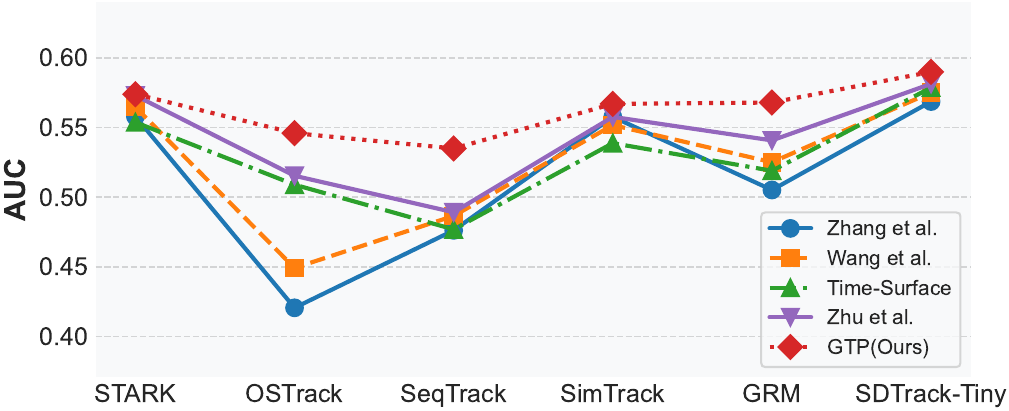}
  \caption{Performance comparison of various event aggregation methods on the FE108 dataset.}
  \label{data_aggregation_auc} 
\vspace{-4pt}
\end{figure}

\textbf{Event Aggregation.} We first evaluate the GTP performance against other event aggregation methods on the FE108 dataset. As shown in Fig.~\ref{data_aggregation_auc}, our method demonstrates significant advantages across five established trackers and the tracker of SDTrack. Notably, several models that previously performed poorly in event-based tracking tasks achieve near-SOTA performance when incorporating GTP. This indicates that the inability of conventional event aggregation methods to provide sufficient temporal information is one of the key factors limiting SOTA SOT models' effectiveness in event-based tracking tasks. This finding also validates the superior performance and generalizability of GTP. Furthermore, the tracker of SDTrack achieves superior accuracy compared to several prominent SOT trackers, regardless of the event aggregation method employed.

\begin{figure}[ht]
  \centering 
  \includegraphics[width=0.47\textwidth]{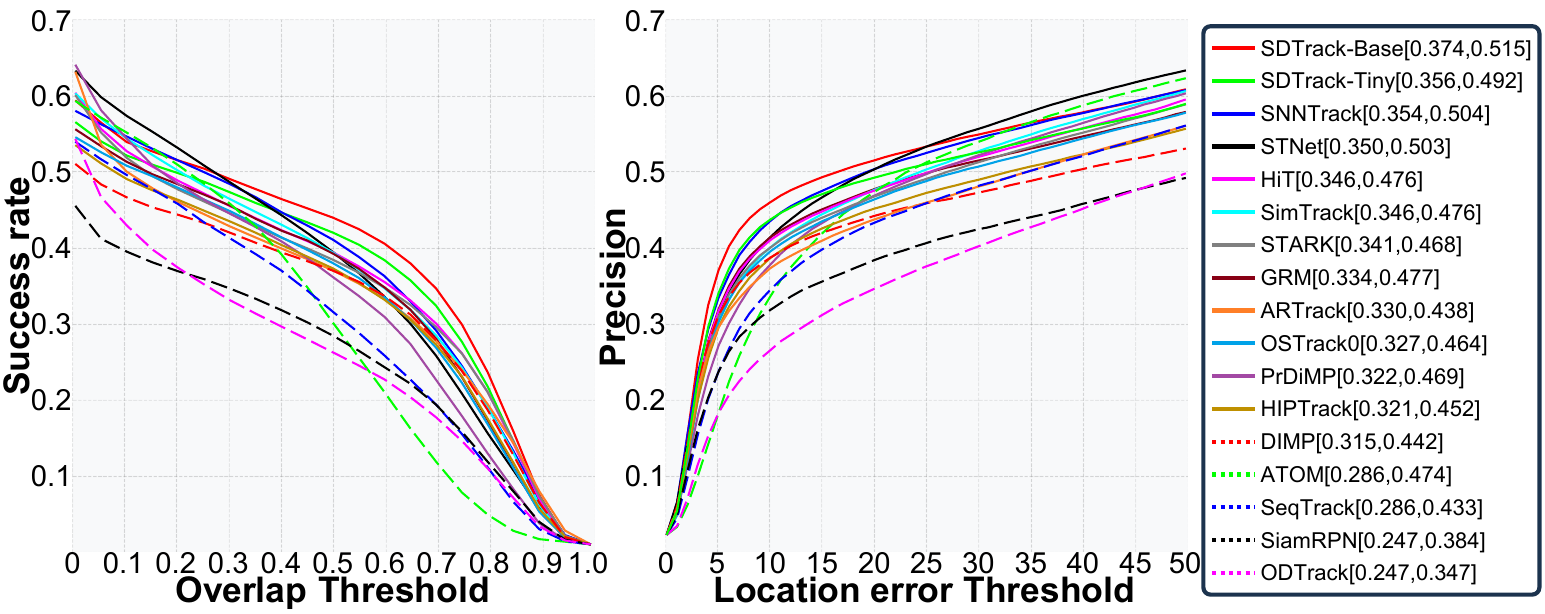}
  \caption{AUC (left) and PR (right) plot on the VisEvent dataset. Best viewed with zooming in.}
  \label{data_processing} 
\end{figure}

\textbf{Event-based Tracking Pipeline.} Based on these findings, SDTrack with different tracker sizes and neurons is evaluated on event-based tracking datasets FE108~\cite{zhang2021object}, VisEvent~\cite{wang2023visevent}, and FELT~\cite{wang2024long}. Detailed specifications are in the Supplementary Material. \textbf{Notably, to highlight the SDTrack tracker's superior performance, most comparison methods employ GTP for event aggregation, significantly improving their performance. Nevertheless, SDTrack still outperforms them.}

As shown in Tab.~\ref{main_table} and Fig.~\ref{data_processing}, across all three datasets, SDTrack-Tiny (SDTrack Pipeline with Tiny-scale tracker) equipped with I-LIF neurons achieves the lowest parameter count and energy consumption. On the FE108 dataset, it improves AUC by 1.6\% and PR by 2.0\% compared to the previous best model, and demonstrates SOTA potential on FELT and VisEvent datasets. The pipelines with Tiny trackers equipped with LIF and temporally-featured I-LIF neurons ($T=2\times2$) show SOTA potential on FE108 and VisEvent datasets, but perform poorly on the long-sequence tracking benchmark FELT dataset. Compared to the Tiny pipeline, the Base pipeline further improves these performance metrics, achieving SOTA results on both AUC and PR metrics across all three event-based tracking datasets.

Notably, while SDTrack-Base maintains a parameter count comparable to previous Base-scale models, it achieves substantially lower theoretical energy consumption. Compared to HiT-B, which was specifically designed for lightweight deployment, our SDTrack-Tiny significantly outperforms across multiple datasets while utilizing less than half the parameters and energy consumption.

\subsection{Ablation Study}
\label{abla}
All ablation experiments follow the same experimental setup and are conducted using SDTrack-Tiny equipped with I-LIF neurons of $T=1\times4$ on FE108.

\begin{figure}[ht]
  \centering 
  \includegraphics[width=0.47\textwidth]{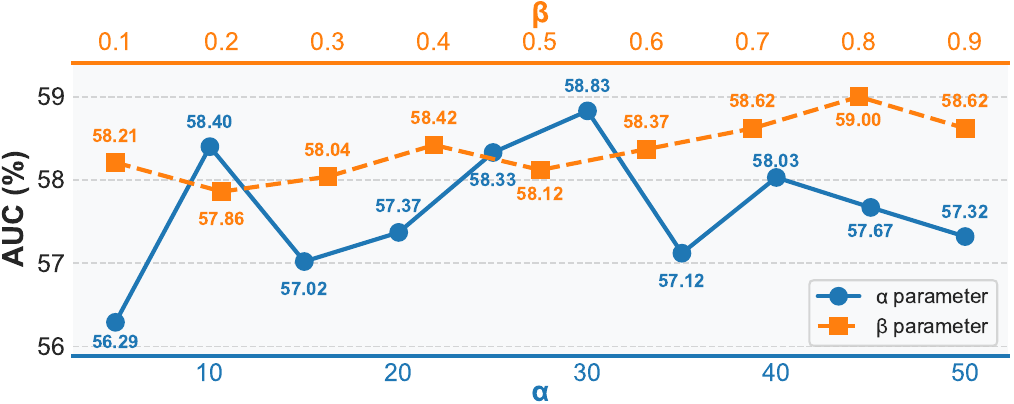}
  \caption{Selection of GTP hyperparameters. During the selection process of $\beta$, $\alpha$ is consistently set to 30.}
  \label{abla_gtp} 
\end{figure}

\textbf{GTP.} The GTP method involves two hyperparameters: $\alpha$ and $\beta$. Through experimental evaluation, we determined that setting $\alpha$ to 30 and $\beta$ to 0.8 is the optimal configuration. The experimental process is illustrated in Fig.~\ref{abla_gtp}.

\textbf{Position-aware.} We quantitatively evaluate the effectiveness of our Intrinsic Position Learning (IPL) approach. As shown in experiment 2 in Tab.~\ref{ablation}, when IPL is not applied and the template and search frames are directly fed into the SNN Conv Module (Siamese network) separately, the PR decreases by 2.04\% due to lack of positional information. This demonstrates that our IPL enables the network to learn positional information effectively. Examining results from experiments 3 and 4 reveals that additional position encoding leads to performance degradation, as the network already naturally learns positional information through the IPL method, and extra positional information introduces noise. Experiments 5-7 demonstrate that the performance improvement stems from learned positional information rather than any form of joint feature extraction. These experiments conclusively demonstrate that our SDTrack does not require explicit position encoding steps, as the necessary positional information is learned inherently.
\begin{table}[]
\centering
\definecolor{purple(x11)}{rgb}{0.63, 0.36, 0.94}
\definecolor{yellow(munsell)}{rgb}{1.0,0.988, 0.957}
\definecolor{green(colorwheel)(x11green)}{rgb}{0.0, 1.0, 0.0}
\definecolor{pink}{rgb}{1.0, 0.85, 0.85}
\setlength{\tabcolsep}{4pt}  
\begin{tabular}{cccc}
\toprule
\# & Method                                & AUC(\%) & PR(\%) \\ \midrule
\rowcolor{gray!15}
\textbf{1}  & \textbf{SDTrack-Tiny}                             & \textbf{59.00}   & \textbf{91.30}  \\
\rowcolor{green!5}
2  & Remove IPL           & 58.10   & 89.66  \\
\rowcolor{green!5}
3  & + Learnable Positional Encoding        & 58.79   & 89.52  \\
\rowcolor{green!5}
4  & + Sinusoidal Positional Encoding       & 58.57   & 90.77  \\
\rowcolor{blue!5}
5  & Intersection Size 0 $\rightarrow$ 32  & 58.22   & 90.14  \\
\rowcolor{blue!5}
6  & Intersection Size 0 $\rightarrow$ 64  & 58.75   & 90.82  \\
\rowcolor{blue!5}
7  & Intersection Size 0 $\rightarrow$ 128 & 43.91   & 73.34  \\
\rowcolor{yellow!5}
8  & Center Head $\rightarrow$ Corner Head & 58.81   & 90.17  \\
\rowcolor{red!5}
9  & No Pretrain & 47.80   & 74.50  \\
\bottomrule
\end{tabular}
\caption{Ablation studies.}
\label{ablation}
\vspace{-8pt}
\end{table}

\textbf{Tracking Head and Pre-training.} Corner and Center Head represent two widely adopted tracking heads in the SOT~\cite{ye2022joint}. We construct SNN versions of both. Experiment 8 in Tab.~\ref{ablation} demonstrates that the Center Head achieves superior performance compared to the Corner Head in SDTrack. Experiment 9 confirms that pre-training in SDTrack, consistent with other trackers, is essential.

\section{Discussion and Limitation}
\label{diss}

\textbf{Comparison with Other Candidate Backbones.} As shown in Tab.~\ref{other_backbone}, our SDTrack, specifically designed for tracking tasks, demonstrates significant performance advantages over other Spike-driven Transformer series networks.

\textbf{Tracking Head Design.} Designing the Decision Layer with spike signals places substantial demands on weight robustness, while floating-point signals necessitate executing this computation on a CPU rather than the neuromorphic chip. As shown in Tab.~\ref{head}, the latter incurs negligible energy overhead, so we treat this design choice as optional.

\begin{table}[]
\centering
\setlength{\tabcolsep}{1.3pt}  
\begin{tabular}{@{}cccc@{}}
\toprule
Backbone   &  Neuron & AUC(\%) & PR(\%) \\ \midrule
Spike-driven Transformer V1~\cite{yao2024spikev1} &  LIF &   52.4  &  85.3 \\
Spike-driven Transformer V2~\cite{yao2024spikev2} &  LIF &   55.9  &  85.8  \\
Spike-driven Transformer V3~\cite{yao2024scaling} & I-LIF &  58.9   & 90.3       \\
\textbf{SDTrack-Tiny (Ours)}   &  \textbf{I-LIF} & \textbf{59.0}  &  \textbf{91.3}  \\ \bottomrule
\end{tabular}
\caption{Comparison with candidate architectures. All settings are aligned with SDTrack.}
\label{other_backbone}
\end{table}

\begin{table}[]
\centering
\begin{tabular}{@{}cccc@{}}
\toprule
 Decision Layer & AUC(\%) & PR(\%) & Power(mJ) \\ \midrule
\textbf{Conv}      & \textbf{59.0}    & \textbf{91.3}   & \textbf{+0.0004}      \\
Spike Neuron~+~Conv & 58.9    & 90.4   & -          \\ \bottomrule
\end{tabular}
\caption{Effect of Decision Layer Design on SDTrack Performance. Experiments are conducted on FE108 using tiny model.}
\label{head}
\end{table}

\begin{figure}[ht]
  \centering 
  \includegraphics[width=0.47\textwidth]{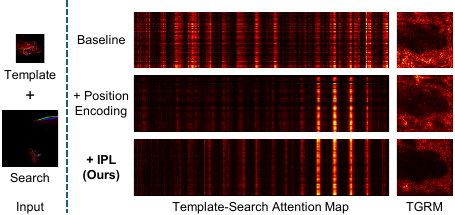}
  \caption{Revealing why IPL outperforms traditional positional encoding. SDTrack-Tiny without IPL serves as baseline. Left: template and search inputs. Right: Attention Map from the last attention module and Template Gradient Response Map (TGRM).}
  \label{heartmap} 
\end{figure}

\textbf{Underlying Mechanisms of IPL Method Effectiveness.} Since tracking tasks are sensitive to positional information, we compare IPL with traditional positional encoding~\cite{ye2022joint,chen2022backbone}. As shown in Fig.~\ref{heartmap}, we visualize Attention Maps and TGRM using the Sec.~\ref{sec:SDTrack} tracker with positional encoding and IPL separately. Each TGRM pixel represents the gradient between the maximum predicted bounding box center score and the corresponding template pixel, effectively measuring tracker understanding and robustness.

Attention Maps show that the baseline struggles with relationship modeling, while both positional encoding and IPL provide positional information to facilitate modeling. TGRM reveals the baseline contains extensive deep red regions, indicating minor target variations cause significant center judgment changes. Positional encoding does not significantly improve this issue, but the IPL method effectively resolves it, demonstrating that IPL improves tracker robustness. The TGRM of the baseline shows obvious red regions corresponding to the target, indicating insufficient understanding of the target's semantic information, which leads to difficulty in locating the target center. After adding positional encoding, the red regions at the target location do not decrease, but after using the IPL method, the red regions at the target location nearly disappear. This demonstrates that the IPL method enhances the tracker's understanding capability of semantic information for tracked targets.

\begin{figure}[ht]
  \centering 
  \includegraphics[width=0.45\textwidth]{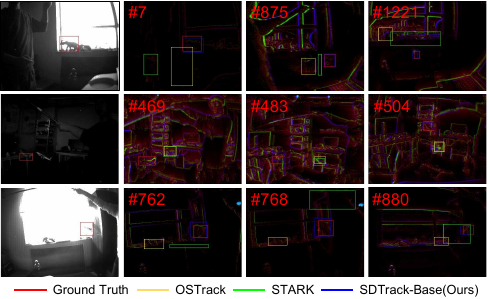}
  \caption{Visualized comparisons of our approach with other excellent trackers OSTrack and STARK during long-sequence tracking. Our method performs better when the target suffers from underexposure, overexposure, and similar object interference.}
  \label{visual_track} 
\end{figure}

\textbf{Performance in Challenging Scenarios.} As shown in Fig.~\ref{visual_track}, SDTrack demonstrates significant advantages in long-sequence tracking and negative conditions through IPL method's enhanced robustness and semantics.

\textbf{Limitation.} This study constructs the first baseline for event-based tracking using entirely SNN, but does not explore multimodal SNN tracking algorithms based on Event+RGB. This will be investigated along with actual deployment on neuromorphic chips in future work.

\section{Conclusion}

In this work, we propose the first transformer-based spike-driven tracking pipeline for event-based tracking, consisting of the GTP method and a spike-driven tracker. GTP provides stronger event representation by recording trajectory information, demonstrating robustness and generalizability across datasets. The tracker learns enhanced positional information and template semantics through IPL without requiring additional positional encoding. Notably, this work fully unleashes the potential of SNNs in event-based tracking tasks. The SDTrack Pipeline operates end-to-end without data augmentation or post-processing, yet achieves performance comparable to state-of-the-art methods while maintaining minimal energy consumption and parameter counts. We hope this work inspires future research.

\section{Acknowledgments} 

This work was supported by the National Natural Science Foundation of China (Grants 62576080 and 62220106008), the Basic Scientific Research Project of Liaoning Provincial Department of Education (Grant JYTMS20230804), and the Sichuan Science and Technology Program (Grant 2024NSFTD0034).

{
    \small
    \bibliographystyle{ieeenat_fullname}
    \bibliography{main}
}

\end{document}